\documentclass[sigconf]{acmart}

\copyrightyear{2021}
\acmYear{2021}
\setcopyright{acmcopyright}
\acmConference[AIES '21] {Proceedings of the 2021 AAAI/ACM Conference on AI, Ethics, and Society}{May 19--21, 2021}{Virtual Event, USA.}
\acmBooktitle{Proceedings of the 2021 AAAI/ACM Conference on AI, Ethics, and Society (AIES '21), May 19--21, 2021, Virtual Event, USA}
\acmPrice{15.00}
\acmISBN{978-1-4503-8473-5/21/05}
\acmDOI{10.1145/3461702.3462554}

\usepackage[inline]{enumitem}
\usepackage{xifthen,graphicx}
\usepackage{mathtools}
\usepackage{amsmath}
\usepackage{amsthm}
\usepackage{bm,mathdots}
\usepackage{makecell}
\usepackage{relsize}
\usepackage{todonotes,comment}
\usepackage{xifthen}
\usepackage{xspace}
\usepackage{stmaryrd}
\usepackage{mathtools}
\usepackage{dsfont}
\usepackage{subcaption}
\usepackage{balance}
\usepackage[capitalise,nameinlink]{cleveref}

\usepackage{pifont}
\usepackage{tablefootnote}
\usepackage{pgfplots}
\usepackage{multirow}

\newtheorem{theorem}{Theorem}

\newtheorem{claim}[theorem]{Claim}

\usepackage{tikz}
\usetikzlibrary{shapes.geometric, arrows}
\tikzstyle{square} = [rectangle, rounded corners, minimum width=1.5cm, minimum height=1cm, text centered, text width=2cm, draw=black, fill=green!30]
\tikzstyle{arrow} = [thick,->,>=stealth]

\newcommand{\figref}[1]{Figure~\ref{fig:#1}}
\newcommand{\figlab}[1]{\label{fig:#1}}

\newcommand{\Server}{\ensuremath{\mathcal{S}}}
\newcommand{\Client}{\ensuremath{\mathcal{C}}}
\newcommand{\Regulator}{\ensuremath{\mathcal{R}}}
\newcommand{\Party}{\ensuremath{\mathcal{P}}}
\newcommand{\Adversary}{\ensuremath{\mathcal{A}}}

\newcommand{\X}{\mathcal{X}}
\newcommand{\Y}{\mathcal{Y}}
\newcommand{\groups}{\mathcal{G}}
\newcommand{\D}{\mathcal{D}}
\newcommand{\M}{\mathcal{M}}
\DeclareMathOperator*{\E}{\mathbb{E}}
\newcommand{\N}{\mathbb{N}}

\newcommand{\Randomize}{\mathtt{aug}}
\newcommand{\indicator}{\mathbb{I}}

\newcommand{\sampsim}[2]{#1 \sim #2}

\newcommand{\samp}[2]{#1 \sim #2}
\newcommand{\yloss}[2]{\ell_{#2}(#1)}
\newcommand{\loss}[2]{\yloss{#1}{#2}}
\newcommand{\yemploss}[2]{\Bar{\ell}_{#2}(#1)}
\newcommand{\emploss}[2]{\yemploss{#1}{#2}}

\newcommand{\likelihood}[2]{{L}_{#2}(#1)}

\newcommand{\Keygen}{\ensuremath{\mathtt{KeyGen}}}
\newcommand{\Sign}{\ensuremath{\mathtt{Sign}}}
\newcommand{\Verify}{\ensuremath{\mathtt{Verify}}}

\newcommand{\certificateId}{\ensuremath{\mathtt{cert}_{ID}}}
\newcommand{\certificateModel}{\ensuremath{\mathtt{cert}_{M}}}

\newcounter{defcounter}

\theoremstyle{plain}

\theoremstyle{definition}
\newtheorem{defn}[defcounter]{Definition}

\newcommand{\FMPC}{\ensuremath{\mathcal{F}_{\mathtt{SC}}}}
\newcommand{\FMcheck}{\ensuremath{\mathcal{F}_{\mathtt{Check}}}}
\newcommand{\PiFramework}{\ensuremath{\pi_{\mathtt{Framework}}}}
\newcommand{\Circuit}{K}

\pgfdeclarelayer{background}
\pgfdeclarelayer{foreground}
\pgfsetlayers{background,main,foreground}

\tikzstyle{Box} = [draw, text width=6em, minimum height=2.5em,align=center, node distance=0.8cm]

\newenvironment{boxfig}[2]{
	\begin{figure}[!th]
		\newcommand{\FigCaption}{#1}
		\newcommand{\FigLabel}{#2}
		
		\begin{center}
			\begin{small}
				\begin{tabular}{@{}|@{~~}l@{~~}|@{}}
					\hline
					\rule[-1.5ex]{0pt}{1ex}
					\begin{minipage}[b]{.925\linewidth}
						\vspace{1ex}
						\smallskip
					}{%
				\end{minipage}\\
				\hline
			\end{tabular}
		\end{small}
		\caption{\FigCaption}
		\figlab{\FigLabel}
	\end{center}
	\vspace{-0.2cm}
\end{figure}
}

\begin{document}
\fancyhead{}
\title{Fairness in the Eyes of the Data: \\ Certifying Machine-Learning Models}
\author{Shahar Segal}
\affiliation{%
  \institution{Tel-Aviv University}
  \city{Tel-Aviv}
  \country{Israel}}
\author{Yossi Adi}
\affiliation{%
  \institution{The Hebrew University of Jerusalem}
  \city{Jerusalem}
  \country{Israel}}
\author{Benny Pinkas}
\affiliation{%
  \institution{Bar-Ilan University}
  \city{Ramat Gan}
  \country{Israel}}
\author{Carsten Baum}
\affiliation{%
  \institution{Aarhus University}
  \city{Aarhus}
  \country{Denmark}}
\author{Chaya Ganesh}
\affiliation{%
  \institution{IISc Bangalore}
  \city{Bangalore}
  \country{India}}
 \author{Joseph Keshet}
\affiliation{%
  \institution{Bar-Ilan University}
  \city{Ramat Gan}
  \country{Israel}}

\begin{abstract}
We present a framework that allows to certify the fairness degree of a model based on an interactive and privacy-preserving test. The framework verifies any trained model, regardless of its training process and architecture. Thus, it allows us to evaluate any deep learning model on multiple fairness definitions empirically. We tackle two scenarios, where either the test data is privately available only to the tester or is publicly known in advance, even to the model creator. We investigate the soundness of the proposed approach using theoretical analysis and present statistical guarantees for the interactive test. Finally, we provide a cryptographic technique to automate fairness testing and certified inference with only black-box access to the model at hand while hiding the participants' sensitive data.
\end{abstract}



\maketitle

\section{Introduction}
\label{sec:introduction}
Machine learning systems are increasingly being used to inform and influence decisions about people, leading to algorithmic outcomes that have powerful personal and societal consequences. For instance, decisions such as \begin{enumerate*}[label=(\roman*)] \item \textit{is an individual likely to commit another crime?}~\cite{compas2016}; or \item \textit{is an individual likely to default on a loan?}~\cite{credit2016} \end{enumerate*} are made using algorithmic predictions. This can be concerning given the many documented cases of models amplifying bias and discrimination from the training data~\cite{buolamwini2018gender,kleinberg2017inherent,corbett2017algorithmic,tatman2017effects}. To address this formally, a line of recent works considers  \textit{fairness} in classification by proposing notions of fairness based on similarity measures and formalizing variants of this notion that provide guarantees against discrimination~\cite{dwork2012fairness,hebert2018multicalibration,kearns2018preventing,kim2019preference}. 



One common scenario in which such a discrimination could potentially happen is a setting with a client and a server. The server classifies queries by a client in an automated way using a machine learning model generated by it. On the other hand, the client wants to make sure its queries are treated fairly and its sensitive data is conserved. If the model itself is not a secret, then a client can potentially run tests (such as the ones implied by the references above) on the model to establish its purported fairness without exposing its data. Making a model public, however, is not always in the interest of the server, since it has invested resources such as expertise, data and computation time for the training -- and therefore often wants the model to remain proprietary. Moreover, sharing models may in some cases raise security or privacy concerns. 
It therefore may be deemed appropriate or necessary to outsource any such test to a semi-trusted third party such as a government entity, which would inspect a model and certify its fairness.
This raises our first question: 


{\em Question 1: Can we design a framework for verifying the fairness of models, giving guarantees to clients while being practically realizable and keeping the model secret from the clients?}

Having such a third party relieves the client from testing fairness, but actually just shifts responsibility to someone who might be more qualified to make a judgement about the model. To minimize the necessary trust between the model owner and the third party, such a test would still be restricted to a black-box scenario.
In addition, constructing such a test for establishing fairness guarantees can be difficult on its own. Given that the sources and amount of data is limited, it might be that the third party can only use data in the fairness test that the model owner is familiar with. This might enable the model owner to design an unfair model which successfully passes the examination by the third party. 

This raises our second question:

{\em Question 2: Can we design a black-box fairness test that gives guarantees even if the test set is (partially) known?}

Here, by a black-box test we mean that a test should only query the model $M$ on different inputs but should not make any assumptions about the actual model parameters.

\paragraph{Our contributions}
In this work, we answer both questions affirmatively. We design an architecture for certifying fairness via verification in machine learning models using three (or more) participants, the model owner (or ``server'') $\Server$, the client $\Client$ and a trusted third party $\Regulator$ (also called ``regulator''). Our architecture uses techniques from cryptography to construct secure protocols for 
\begin{enumerate}[leftmargin=*]
    \item An interactive test between $\Server$ and $\Regulator$ allowing to verify with high probability that a model $M$ provided by $\Server$ is fair with respect to a set of pre-defined groups. While ensuring that $\Regulator$ does not learn $M$. This test considers scenarios where $\Server$ {\em is} or {\em is not} aware of the test data. $\Regulator$ is not involved in the training of $M$, it only performs certification.
    \item An interactive computation between $\Server$ and $\Client$ which computes a prediction $\hat{y}=M(x)$ from an input $x$ and a model $M$. The interactive computation neither leaks $M$ to $\Client$ nor $x$ to $\Server$, and yet makes sure that the model that was used in the prediction has been certified by $\Regulator$ beforehand.
\end{enumerate}
Our work provides fairness tests necessary for these protocols that have black-box to the model and uses existing highly efficient cryptographic primitives to implement the tests securely. While we motivate the underlying ideas of these tests on an intuitive level and give formal arguments for their soundness, we also provide experimental evidence that the hypotheses that make our tests possible are viable. Since secure and privacy-preserving  computation of models, for both training and inference, is a very active research area~(e.g. \citet{secureml,gazelle,chameleon,cryptflow,van2020trade,guo2020secure}), the performance of the current solutions in this field is continuously improving. As our work assumes the existence of secure protocols for inference, and investigates how to add fairness on top of these in a generic way that is independent of the underlying training algorithm, our approach will benefit in practicality from any independent progress that is made in this direction. 

\paragraph{Related work}
Fairness in algorithms was first investigated by~\citet{friedman1996bias}. Since then, further research into data as a source of unfairness in ML decisions has been done~\cite{kay2015unequal,caliskan2017semantics}. \citet{quantitative_verification} showed how to verify properties of a DNN (fairness among them). In their work they encode the network into \emph{Conjunctive Normal Forms} and then test if it will likely fulfill certain logical constraints. In comparison, our approach is independent of the concrete model parameters and architecture.

Several statistical measures of unfairness, and fairness criteria are studied by~\citet{feldman2015certifying,zemel2013learning}.
These and subsequent works achieve statistical notions of fairness through
post-processing the training data, and/or by
enforcing constraints at training time. Our work differs from this line of research in that we want to guarantee fairness which is enforced obliviously of the training process.
\citet{dwork2012fairness} shows that  statistical notions of fairness are inadequate, while~\citet{corbett2017algorithmic} established that model calibration does not rule out unfair decisions. These results emphasize that fairness is nuanced, complicated, application-specific, and can depend on legal and social contexts. In this work, we answer the orthogonal question of designing a fairness test for any machine learning model, \textit{given} an accepted fairness definition.

Most relevant prior work to ours is the study by~\citet{blind_justice}. In this research the authors suggest to use cryptographic primatives for \emph{fairness certification}, \emph{fair model training}, and \emph{model decision verification}. However, this study was mainly focused on certifying a model via fair training. Additionally it did not provide analysis and guarantees for model fairness certification. Our study focuses on certifying fairness via verification of any existing machine learning models, regardless of the training process. Since the framework is oblivious to the training process, multiple fairness definitions can be certified post-training, even if the training process did not take them into account. We analyze our framework from both theoretical and practical points of view while providing guarantees based on the number of samples available in the test set. We also explore a different scenario where these samples are known to $\Server$ during model training, which makes the certification harder. 

\section{Preliminaries}
\label{sec:preliminaries}
Let $\X$ be the set of possible inputs, $\groups$ be a finite set of groups that are relevant for fairness (e.g., ethnic groups) and $\Y$ be a finite set of labels.
We suppose $\X \times \groups \times \Y$ is drawn from a probability space $\Omega$ with an unknown distribution $\D$. Let $M$ be a trained model for a classification task of $\D$, we denote $M(x)$ for classification of input $x \in \X$.

\begin{figure*}[t!]
    \centering
    \begin{subfigure}{.4\textwidth}
        \begin{center}
        
            \begin{tikzpicture}[scale = 0.6,transform shape,box/.style={draw,rounded corners,align=center},  thick]
            \node[Box, fill=white] (Sign) {\Large $\Sign$};
            \node[inner sep=0,minimum size=0, left=1.5cm of Sign] (SupNode) {};
            \node[Box, below right=1.0cm and 2.0cm of Sign, fill=white] (Verify) {\large $\Verify$};
            \node[Box, above left=1cm and 1cm of Sign, fill=white] (Keygen) {\large $\Keygen$};
            
            \draw[->] (SupNode)+(-2cm,0cm) -- node[anchor=south] {$m$} (SupNode) -- (Sign.west);
            \draw[->] (SupNode) |- ([yshift=-0.2cm]Verify.west);
            
            \draw[->] (Keygen.east)+(0,-0.2) -| node[anchor=west] {$sk$} (Sign.north);
            
            \draw[->] (Keygen.east)+(0,0.2) -| node[anchor=west] {$vk$} (Verify.north);

            \draw[->] (Sign.east) -- +(1,0) |- node[anchor=south west] {$\sigma$} ([yshift=0.2cm]Verify.west);

            \draw[->] (Verify.east) -- node[anchor=south] {$0/1$} +(2.0,0);
            
            \end{tikzpicture}
        \end{center}
    \end{subfigure}%
    \begin{subfigure}{.4\textwidth}
        \begin{center}
            \begin{tikzpicture}[scale = 0.6 ,transform shape,box/.style={draw,rounded corners,align=center},  thick]
               \node (bigtriag-lt) at (-5.0,  5.0) {}; 
               \node (bigtriag-rt) at (5.0, 5.0) {}; 
               \node (bigtriag-cb) at ( 0.0, 0.0) {}; 
               
               \node (finp-lt) at (-4,  4.6) {}; 
               \node (finp-rt) at (-2, 4.6) {}; 
               \node (finp-cb) at ( -3, 3.6) {}; 
               
               \node (sinp-lt) at (-1.5,  4.6) {}; 
               \node (sinp-rt) at (0.5, 4.6) {}; 
               \node (sinp-cb) at ( -0.5, 3.6) {}; 
               
               \node (linp-lt) at (2,  4.6) {}; 
               \node (linp-rt) at (4, 4.6) {}; 
               \node (linp-cb) at ( 3, 3.6) {};

               \node (fcomb-lt) at (-2.75,  3.2) {}; 
               \node (fcomb-rt) at (-0.75, 3.2) {}; 
               \node (fcomb-cb) at ( -1.75, 2.2) {};

               \node (last-lt) at (-1.0,  1.4) {}; 
               \node (last-rt) at (1.0, 1.4) {}; 
               \node (last-cb) at ( 0, 0.4) {};

            \filldraw[draw=black, fill=gray!20] (bigtriag-lt.center)--(bigtriag-rt.center)--(bigtriag-cb.center)--(bigtriag-lt.center);
            
            \filldraw[draw=black, fill=white] (finp-lt.center)--(finp-rt.center)--(finp-cb.center)--(finp-lt.center);
            \filldraw[draw=black, fill=white] (sinp-lt.center)--(sinp-rt.center)--(sinp-cb.center)--(sinp-lt.center);
            \filldraw[draw=black, fill=white] (fcomb-lt.center)--(fcomb-rt.center)--(fcomb-cb.center)--(fcomb-lt.center);
            \filldraw[draw=black, fill=white] (last-lt.center)--(last-rt.center)--(last-cb.center)--(last-lt.center);
            
            \filldraw[draw=black, fill=white] (linp-lt.center)--(linp-rt.center)--(linp-cb.center)--(linp-lt.center);

            \node (bigtriag-caption) [below=0.3 of bigtriag-cb] {$\widehat{H}_k(m_1,\dots,m_\ell)$};
            
            \node (finp-firstinput-caption) [above right=0.8 and 0.04 of finp-lt] {$m_1$};
            \node (finp-secondinput-caption) [above left=0.8 and 0.04 of finp-rt] {$m_2$};
            \node (sinp-firstinput-caption) [above right=0.8 and 0.04 of sinp-lt] {$m_3$};
            \node (sinp-secondinput-caption) [above left=0.8 and 0.04 of sinp-rt] {$m_4$};
            \node (linp-firstinput-caption) [above right=0.8 and 0.04 of linp-lt] {$m_{\ell-1}$};
            \node (linp-secondinput-caption) [above left=0.8 and 0.04 of linp-rt] {$m_\ell$};

            \node (dots-left-caption) [below right=0.0 and 0.2 of fcomb-cb] {$\dots$};
            \node (dots-right-caption) [above right=0.25 and -0.25 of last-rt] {$\iddots$};
            
            \node (last-caption) [above=0.25 of last-cb] {$H_k(\cdot)$};
            \node (first-caption) [above=0.25 of finp-cb] {$H_k(\cdot)$};
            \node (second-caption) [above=0.25 of sinp-cb] {$H_k(\cdot)$};
            \node (last-input-caption) [above=0.25 of linp-cb] {$H_k(\cdot)$};
            \node (combination-caption) [above=0.25 of fcomb-cb] {$H_k(\cdot)$};
            
            \node (dots-triangles-first-row) [right=0.9 of second-caption] {$\dots$};
            \node (dots-triangles-second-row) [right=0.9 of combination-caption] {$\dots$};
            
            \draw[densely dotted] (last-cb.center)--(bigtriag-cb.center);
            \draw[-] (bigtriag-cb.center)--(bigtriag-caption) ;
            \draw[densely dotted] ([xshift=0.5cm]finp-lt.center)--([xshift=0.5cm,yshift=0.4cm]finp-lt.center);
            \draw[-] (finp-firstinput-caption)--([xshift=0.5cm,yshift=0.4cm]finp-lt.center);
            \draw[densely dotted] ([xshift=-0.5cm]finp-rt.center)--([xshift=-0.5cm,yshift=0.4cm]finp-rt.center);
            \draw[-] (finp-secondinput-caption)--([xshift=-0.5cm,yshift=0.4cm]finp-rt.center);
            \draw[densely dotted] ([xshift=0.5cm]sinp-lt.center)--([xshift=0.5cm,yshift=0.4cm]sinp-lt.center);
            \draw[-] (sinp-firstinput-caption)--([xshift=0.5cm,yshift=0.4cm]sinp-lt.center);
            \draw[densely dotted] ([xshift=-0.5cm]sinp-rt.center)--([xshift=-0.5cm,yshift=0.4cm]sinp-rt.center);
            \draw[-] (sinp-secondinput-caption)--([xshift=-0.5cm,yshift=0.4cm]sinp-rt.center);

            \draw[densely dotted] ([xshift=0.5cm]linp-lt.center)--([xshift=0.5cm,yshift=0.4cm]linp-lt.center);
            \draw[-] (linp-firstinput-caption)+(-0.15cm,-0.25cm)--([xshift=0.5cm,yshift=0.4cm]linp-lt.center);
            \draw[densely dotted] ([xshift=-0.5cm]linp-rt.center)--([xshift=-0.5cm,yshift=0.4cm]linp-rt.center);
            \draw[-] (linp-secondinput-caption)--([xshift=-0.5cm,yshift=0.4cm]linp-rt.center);
            
            
            \draw[-] (finp-cb.center)--([yshift=-0.2cm]finp-cb.center)|-([xshift=0.5cm,yshift=0.2cm] fcomb-lt.center)--([xshift=0.5cm] fcomb-lt.center);
            \draw[-] (sinp-cb.center)--([yshift=-0.2cm]sinp-cb.center)|-([xshift=-0.5cm,yshift=0.2cm] fcomb-rt.center)--([xshift=-0.5cm] fcomb-rt.center);
            
            \draw[dashed, color=gray] (fcomb-cb.center) -- ([yshift=0.4] fcomb-cb.center) |-(dots-left-caption);
            \draw[dashed, color=gray] ([xshift=0.5cm]last-lt.center) -- ([yshift=0.4cm, xshift=0.5cm] last-lt.center) |-(dots-left-caption);
            
            \draw[dashed, color=gray] (linp-cb.center) -- ([yshift=-0.4] linp-cb.center) |-([xshift=-1.6cm, yshift=-0.4cm]linp-cb.center) |- ([xshift=-1.6cm, yshift=-1.2cm]linp-cb.center);
            \draw[dashed, color=gray] ([xshift=-0.5cm]last-rt.center) -- ([yshift=0.5cm, xshift=-0.5cm] last-rt.center) |-([yshift=0.5cm, xshift=0.0cm] last-rt.center);
            
            \end{tikzpicture}
        \end{center}
    \end{subfigure}
    \caption{\footnotesize Signatures (left), Merkle Trees (right)}
    \label{fig:signatures and merkle trees}
    \vspace{-0.2cm}
\end{figure*}
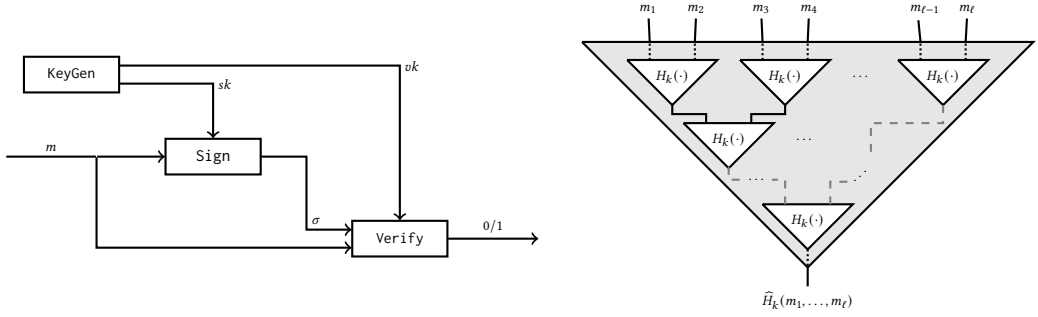


The goal of training a model $M$ is usually to achieve low error on unseen data. In addition, when dealing with model fairness, we also take into account a measurement with respect to $\groups$. While there are plenty of fairness measurements~\cite{verma2018fairness}, here we focus on group risk and likelihood based definitions, specifically: \emph{overall risk equality} (ORE), \emph{equalized odds} (EO) and \emph{demographic parity} (DP). 
First, we define the conditional \emph{risk} and \emph{likelihood} respectively:
\begin{align}
     \tag{Risk} \loss{M}{g} = ~&\E_{\sampsim{(x,g',y')}{\D}} \left[ \indicator\left\{M(x) \neq y' \right\} | g'=g \right]
     \\
    \tag{Risk with label condition} \loss{M}{g,y} = ~&\E_{\sampsim{(x,g',y')}{\D}} \left[ \indicator\left\{M(x) \neq y'\right\} | g'=g, y'=y \right]
    \\
    \tag{Likelihood} \likelihood{M}{g,y}  = ~&\E_{\sampsim{(x,g',y')}{\D}} \left[ \indicator\left\{M(x) = y\right\} | g'=g \right]
\end{align}
where $\indicator\{\pi\}$ is an indicator function with a predicate $\pi$. 
The empirical conditional risk is defined for a given independent sample set $\samp{T=\{(x_1,g_1,y_1),...,(x_m,g_my_m)\}}{\D^m}$ as: 
\begin{align}
    \label{eq:emp_group_risk}
    \emploss{M,T}{g} &= \frac{1}{m_g}\sum_{i=1}^{m} \indicator\{M(x_i) \neq y_i \wedge g_i=g\}
\end{align}
where $m_g$ is the number of samples in $T$ from group $g$. 

We define a metric called the \emph{fairness gap} to be the maximal margin between any two groups (and labels). Formally, we use three well-known measurements:
\begin{align}
    \tag{ORE} &~~~\max_{g_0,g_1 \in \groups} ~~~~|\loss{M}{g_0} - \loss{M}{g_1}|
    \\
    \tag{EO} &\max_{g_0,g_1 \in \groups, y \in \Y} |\loss{M}{g_0, y} - \loss{M}{g_1, y}|
    \\
    \tag{DP} &\max_{g_0,g_1 \in \groups, y \in \Y} |\likelihood{M}{g_0, y} - \likelihood{M}{g_1, y}|
\end{align}
Likewise, the \emph{empirical fairness gap} (EFG) is defined using the empirical approximation of each measurement respectively.

Lastly, we call model $M$ \textbf{$\epsilon$-fair} on $(\groups, \D)$ with respect to a fairness measurement, if its fairness gap is smaller than $\epsilon$ with confidence $1-\delta$, 
which is similar to PAC-style fairness~\cite{rothblum2018probably}. A model $M$ is then called \textbf{$\epsilon$-fair} on $(\groups, \D)$ under the ORE metric if:  
\begin{equation}
        \label{eq:epsfair}
        \Pr \Big[\max_{g_0,g_1 \in \groups} |\loss{M}{g_0} - \loss{M}{g_1}| > \epsilon \Big] \leq \delta
\end{equation}
Plugging-in the fairness gap metric for EO and DP yields the corresponding $\epsilon$-fairness definitions.

\section{Cryptographic Primitives}

\label{sec:cryptoprim}
We now describe the cryptographic primitives that are necessary to implement the proposed framework in more detail: \emph{Signatures}, \emph{Collision-Resistant Hash Functions} and \emph{Secure Computation}. 

\textbf{Signatures.} 
Cryptographic signatures can be thought of as a computational analogue to hand-written signatures.
We give a schematic explanation of signature schemes in Figure \ref{fig:signatures and merkle trees}. Here, a pair of a public verification key $vk$ and a secret signing key $sk$ are generated together by the key generation algorithm $\Keygen$. $sk$ will be used by the signing algorithm $\Sign$ to create a signature $\sigma$ on a message $m$, while the verification algorithm $\Verify$ decides if a pair $(m,\sigma)$ is valid according to the verification key $vk$ or not. Secure signature schemes guarantee \emph{unforgeability}, which means that given $vk$ and arbitrarily many signature  pairs $\{(m_i, \sigma_i)\}_{i\in[\ell]}$, it is hard to generate a valid signature $\sigma$ on a message $m$, where $m\neq m_i$ for all $i\in [\ell]$.

\textbf{Collision-Resistant Hashing.}
We will use a \emph{Collision-Resistant Hash Function} $H_k: \{0,1\}^n \times \{0,1\}^{2n}\rightarrow \{0,1\}^n$, which is an efficiently computable function such that it is hard for any polynomial-time algorithm (in $n$) that is given a random $k$ to come up with $x_1,x_2$ such that $H_k(x_1)=H_k(x_2)$.  In practice, one uses e.g. SHA-3 to implement $H_k$ for a $k$ that is fixed in advance. Since the input length of SHA-3 is fixed, in order to hash longer messages, one can apply $H_k$ recursively using a Merkle Tree (see \figref{signatures and merkle trees}). For such a Merkle Tree it can be proven that if $H_k$ is collision-resistant then $\hat{H}_k$ is too.

\textbf{Secure Computation.}
We further let parties perform computations on shared data such that the computation does not reveal their inputs, for purposes as mentioned in the next Section.

Secure Computation can be imagined as the existence of a ``trusted third party'' $\FMPC$ which performs a computational task for certain parties. $\FMPC$ would receive the inputs from both participants, do the computation, and send the output to the participants.
The task of this party is  outlined in \figref{securecomp}. As is common in the secure computation literature,  this description assumes that the computation is done by a circuit $\Circuit$.   Participant $\Party_1$ provides to the trusted party its input $x_1$, while participant $\Party_2$ provides its input $x_2$. The trusted party computes $\Circuit(x_1,x_2)$ and sends its outputs to the respective participants. By this definition  this ``idealized box'' $\FMPC$ achieves the desired privacy objective.

\begin{boxfig}{ \footnotesize A Trusted Third Party $\FMPC$ for Secure Computation.}{securecomp}
	Two parties $\Party_1, \Party_2$ can talk to this trusted third party. 
	\begin{description}
		\item[Input:] Upon message $(\mathtt{Input}-\Party_1,x_1)$ from $\Party_1$ and $(\mathtt{Input}-\Party_2,x_2)$ from $\Party_2$ store $x_1,x_2$ locally. 
		\item[Compute:] Upon input $(\mathtt{Compute}, \Circuit)$ from $\Party_1$ and $\Party_2$ and if $x_1,x_2$ have been stored:
		\begin{enumerate}
		    \item Check if $x_1,x_2$ have suitable size for the circuit $\Circuit$. If not, output $(\mathtt{Abort})$.
		    \item If $x_1, x_2$ have suitable size then	compute $(y_1, y_2)=\Circuit(x_1, x_2)$ and store $y_1, y_2$ locally.
		\end{enumerate}
	
		\item[Output:] Upon input $(\mathtt{Output})$ from $\Party_1$ and $\Party_2$ and if $y_1, y_2$ have been computed, send $y_1$ to $\Party_1$ and $y_2$ to $\Party_2$.
	\end{description}
\end{boxfig}
Such a trusted third party $\FMPC$ as described in \figref{securecomp} does not necessarily exist in the real world, but \emph{it can be emulated} using cryptographic tools as a protocol consisting of two (or more) entities sending messages to each other over a network. Guarantees in these protocols can be given if at least one of the participants is acting honestly throughout the process. 

The two most popular approaches for implementing \figref{securecomp} are based on cryptographic paradigms called \emph{Fully Homomorphic Encryption} (FHE) and \emph{Secure Multiparty Computation} (MPC). For comparison, current FHE schemes are constrained by their demand for computational power and they at best can evaluate a few hundred AND-gates of the circuit $\Circuit$ per second. MPC on the other hand, which has a higher demand in terms of communication,  can achieve a much better throughput. In particular, there exist MPC schemes that are tailored at efficiently implementing the function $M(\cdot)$~\footnote{A recent framework for privacy preserving machine learning using MPC built on PyTorch: CrypTen 
\cite{CrypTen}.
}~\cite{secure_eval_quantized,secure_eval_rings}.

\section{The Framework}
\label{sec:framework}
We present our framework from a broad overview, and leave most of the implementation details to later sections. There we describe two interactive tests to verify fairness and more wholesome view on the cryptographic aspects. For now we focus on the general flow and interaction between the different participants we previously described. We also make their roles more explicit and describe the security guarantees that are given to each of them as well as the trust relations. Note that we discuss our framework with respect to three participants but it can easily be generalized to any larger number. In particular, it allows for a large number of regulators $\{\Regulator_i\}_{i=1}^k$ that a client $\Client$ can choose from or even perform the regulators role by itself. 

\begin{itemize}[leftmargin=*]
\item The \emph{Server $\Server$} initially generates the model $M$. Its main objective is to keep $M$ secret. He may try to use an unfair model towards $\Regulator$ or $\Client$.
\item The \emph{Client $\Client$} has a private input $x$ and wishes to obtain $\hat{y} \gets M(x)$, where the server provides $M$. 
The objective of $\Client$ is to ensure that $M$ is fair while keeping $x$ private.
\item  The third participant is the \emph{Regulator $\Regulator$} who should neither learn $M$ nor $x$ or $y$. After $\Server$ proves the fairness of model $M$, $\Regulator$ outputs certificate $\certificateModel$ for the model to attest its validity. $\certificateModel$ is tied to another certificate $\Regulator$ issued, $\certificateId$, which serves as the identity of $\Regulator$. 
When $\certificateModel$ is shown to $\Client$ it can verify that indeed $\Regulator$ certified the model using $\certificateId$. In addition, $\Regulator$ has access to the sample set $T$ in order to check fairness, which is possibly known to $\Server$.
\end{itemize}

In terms of modeling security we assume that $\Server$ will try by any means to get an unfair model certified, use an unfair model to a client or learn $\Client$'s input. In particular, $\Server$ may actively deviate from any specified program or protocol in any way. On the other side, we consider that $\Client$ and $\Regulator$ would follow the protocol but may try to learn information about $\Server$'s model in the process. In more cryptographic terms we consider a  malicious $\Server$ while $\Client$ and $\Regulator$ are semi-honest. We further assume that either party is computationally polynomially time-bounded.

The certificates $\certificateId,\certificateModel$ are implemented using a digital signature scheme and a collision-resistant hash function. Roughly, $\certificateId$ is a public verification key that is tied to the identity of $\Regulator$, and $\certificateModel$ is a signature on a  compressed version of the model that is computed using a collision-resistant hash function. Here, a cryptographic signature ensures that only $\Regulator$ could issue $\certificateModel$, while the hash function forces $\Server$ to use the same model with $\Client$ that he used when obtaining $\certificateModel$.

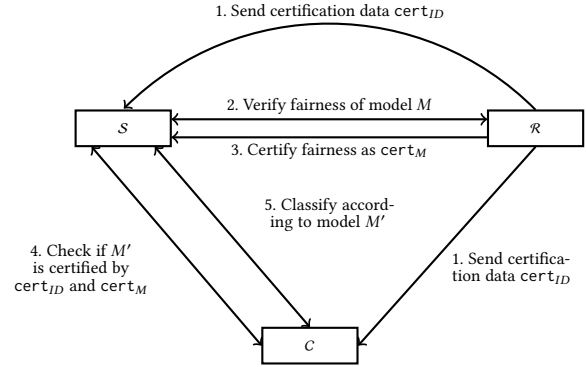
\begin{figure}[t!]
    \begin{center}
            \begin{tikzpicture}[scale = 0.6, transform shape,box/.style={draw,rounded corners,align=center},  thick]
            \node[Box, fill=white] (Server) {$\Server$};
            \node[Box, below right=4.0cm and 2.0cm of Server, fill=white] (Client) {$\Client$};
            \node[Box, right=7.0cm of Server, fill=white] (Regulator) {$\Regulator$};
            
            \node[inner sep=0,minimum size=0, above=0cm of Server] {} edge[<-, bend left=45] node[auto] {\Large 1. Send certification data $\certificateId$} (Regulator.north);
            \draw[->] (Regulator.south) -- node[auto, text width=3cm] {\Large 1. Send certification data $\certificateId$} (Client.east);
            
            \draw[<->] ([yshift=0.2cm]Server.east) -- node[auto] {\Large 2. Verify fairness of model $M$} ([yshift=0.2cm]Regulator.west);
            
            \draw[<-] ([yshift=-0.2cm]Server.east) -- node[anchor=north] {\Large 3. Certify fairness as $\certificateModel$} ([yshift=-0.2cm]Regulator.west);
            
            \draw[<->] ([xshift=0.7cm]Server.south) -- node[auto, text centered, text width=4cm] {\Large 5. Classify according  to model $M'$} (Client.north);
            \draw[<->] ([xshift=-0.7cm]Server.south) -- node[anchor=east, text centered, text width=3cm, yshift=-0.5cm, xshift=-0.5cm] {\Large 4. Check if $M'$ is certified by $\certificateId$ and $\certificateModel$} (Client.west);
        \end{tikzpicture}
    \end{center}
    \caption{
    \small
    Certification and Verification}
    \label{fig:registration_and_verification}
\end{figure}

At the beginning of the protocol, $\Regulator$ generates its public certificate $\certificateId$ and makes it available. $\Server$ and $\Regulator$ then interact to generate the model certificate $\certificateModel$ for model $M$. In the process $\Regulator$ is allowed to query $M$ an arbitrary number of times to ensure fairness. To perform an inference by $\Server$ and $\Client$, both first agree on a regulator certificate $\certificateId$ that they will use. Then, $\Client$ obtains an output $\hat{y}$ based on its input $x$ and on a model $M'$ provided\footnote{Here we write $M'$ to denote that technically a malicious $\Server$ could try to perform the inference with whatever model $M'$ it wants. The job of the inference algorithm is then to enforce that $M'=M$ for some previously certified $M$.} by $\Server$. Here, $\Client$ only accepts $\hat{y}$ if $M'$ is certified by the regulator behind $\certificateId$ for fairness. $\Client$ does not learn anything about $M'$, besides $\hat{y}$. Fig. \ref{fig:registration_and_verification} describes the aforementioned process schematically. 


Both the inference on $M$ and the verification of $\certificateModel$ that are necessary in Fig.~\ref{fig:registration_and_verification} could be done easily if $\Server,\Regulator$ and $\Client$ would have access to a ``trusted third party'' $\FMPC$  which performs the computational task for them. That trusted third party would receive inputs from  participants, do the computation, and send the output back to them. In our case, $\FMPC$ would receive all secret input from the participants and send $\certificateModel$ to $\Server$ after verifying $M$ is fair. $\FMPC$ can also send $M(x)$ to $\Client$ if model $M'$ is certified in that manner.
As mentioned in the previous section, such a "trusted third party" can be emulated using cryptographic protocols for Secure Computation. The use of cryptography guarantees that the protocol is secure against a \emph{malicious} $\Server$ or semi-honest $\Client,\Regulator$.





\section{Verifying Fairness Interactively}\label{sec:fairness}

We now turn to introduce two interactive tests which allow $\Regulator$ to determine if a model $M$ is $\epsilon$-fair:
\begin{enumerate}[leftmargin=*]
    \item The model $M$ is queried using a sample set $T$ which is \emph{unknown} to $\Server$. We show that fairness guarantees about $M$ can be made given by the empirical fairness gap (EFG) and a lower-bound on the minimal size of the set $T$ with respect to each group $g\in \groups$.
    \item The model $M$ is queried using a sample set $\tilde{T}$ which is derived from a set $T$ in an augmented and randomized fashion. 
    The set $T$ as well as the augmentation algorithm are known to $\Server$ in advance. We show that this test implies $\epsilon$-fairness of $M$, given that the augmentation impacts fair and unfair models in a different way.
\end{enumerate}
Both of the aforementioned tests are independent of the representation of $M$, make no requirement on its training algorithm and only require access to $M(\cdot)$ for different inputs. 
All claims below are based on $\epsilon$-fairness with confidence $1-\delta$ under the ORE fairness metric, however these can be modified to other group-based fairness definitions.

\paragraph{Verifying Fairness using Private Data.}
\label{sec:private_data_test}
The simpler case is when the model $M$ is queried using an i.i.d sample set $T$ which is \emph{unknown} to $\Server$. This setup is rather standard in machine learning, as verification using i.i.d samples can be achieved via classic concentration bounds from PAC learning~\cite{valiant1984theory,vapnik2006estimation}. We apply similar bounds to assess the minimum number of samples needed for fairness verification.


Denote the \emph{Empirical Fairness Gap} as 
$$EFG = \max_{g_0,g_1 \in \groups} |\emploss{M, T}{g_0} - \emploss{M, T}{g_1}|.$$ 
The following states the conditions which guarantee that a model is $\epsilon$-fair with a confidence $1-\delta$.  

\begin{claim}
\label{thm:sampling}
 A model $M$ is $\epsilon$-fair with confidence $1-\delta$ if:
    \begin{equation}
    \label{eq:minbound-fairness-private}
        EFG < \epsilon ~~~~\text{and}~~~~\min_{g \in \groups}{m_{g}} \geq \frac{2}{(\epsilon-EFG)^2}\ln{\frac{2|\groups||\Y|}{\delta}}
    \end{equation}
for $\samp{T=\{(x_1,g_1,y_1),...,(x_m,g_m,y_m)\}}{\D^m}$, where $m_g$, as in Eq.~$(\ref{eq:emp_group_risk})$, denotes the number of occurrences of $g$ in $T$.
\end{claim}

DP and EO can be achieved by using the corresponding EFG definition and minimizing over $\groups \times \Y$, counting $m_g$ and $m_{g,y}$ respectively in (\ref{eq:minbound-fairness-private}). The full proof of the above claim can be found in the supplementary materials.

\paragraph{Verifying Fairness using Augmented Data.}
\label{sec:public data test}

The disadvantage of the aforementioned test is that all test data $T$ must be hidden so that the model generator $\Server$ cannot use it to adapt $M$ accordingly. In other settings, we would like to test for fairness using public data, which can be known to $\Server$. This setting is realistic in many scenarios. For example, if labelled data is costly, getting unique labelled data for a test will be difficult for $\Regulator$.

A straightforward argument against this approach is once the data is publicly available, $T$ is not chosen independently of $M$. Thus, a malicious $\Server$ can create an unfair model that memorizes the set $T$ and responds fairly on it, so that it passes the test outlined above. To counter such dishonest training, we need a method to alter the existing samples and force some sort of generalization abilities. We therefore define the notion of an \emph{augmentor}. An augmentor applies random augmentations to the input which alter the sample but still preserves its label and group with high probability. Here we use it to generate \emph{new} samples for querying that with high probability were not seen during model training. This is a necessary but not sufficient condition in order to ensure a valid test for $M$. For example, consider an augmentor that only alters the first few pixels of the image. A model that simply ignores those pixels can still overfit on the rest of the image and pass any test.

Hence, we suggest to use a set of randomized augmentation functions to reduce memorization capabilities of an adversary. For this, the assumption is that $\epsilon$-fair models behave differently from unfair models when queried against the samples augmented by the augmentor. Then, this different behavior can be leveraged to expose the unfair nature of certain models. Our approach follows this assumption to construct a querying test set in the same fashion of the test that was previously defined. 

More specifically, define an algorithm augmentor $\Randomize: \X \times \groups \times \{0,1\}^\tau \to \X$ that gets as input a random string and a sample and outputs a new augmented sample. The label and group of the new sample should be the same as the original sample with high probability. 

We re-define the conditional risk to be on an augmented sample from $\D$. Formally:
\begin{align*}
    \loss{M}{g, \Randomize} = \E_{\sampsim{(x,g,y)}{\D}, r\gets \{0,1\}^\tau} 
    & \Big[  \indicator\{M(\Randomize(x,g;r)) \neq y\} ~\Big].
\end{align*}
As mentioned before, there is no guarantee that samples augmented by $\Randomize$ yield better results than $T$ itself. We need an additional assumption on the behavior of fair and unfair models when shown augmented samples from $\Randomize$, thus we call a class of models $\M$ \emph{detectable} if it fulfills that assumption.

\begin{defn}[$(\epsilon, \alpha, \Randomize)-$detectable fairness]\label{def:detectable_fairness}

Let $\Adversary$ be an arbitrary training algorithm which outputs a model in $\M$. $\M$ has $(\epsilon, \alpha, \Randomize)$-detectable fairness on $\D$ if there exists $m \in \N$ such that for any $\samp{T}{\D^m}$ and $\samp{M}{\Adversary(\D,T,\Randomize,\alpha)}$, $M$ is $\epsilon$-fair if:
    $$\max_{g_0, g_1 \in \groups} |\loss{M}{g_0, \Randomize} - \loss{M}{g_1, \Randomize}| \leq \alpha~.$$
\end{defn}



\cref{def:detectable_fairness} allows us to build an interactive test and to empirically find parameters $\epsilon, \alpha, m$ and an augmentor for which it appears to be true. The intuition behind \cref{def:detectable_fairness} is that in order to cheat the test, M is required to behave differently on the augmented data than on the new unobserved samples. In practice, augmentations are commonly used to improve robustness and generalization, thus M is less likely to be able to generalize on a large set of them without the risk of exposing its unfair nature, as $\Randomize$ challenges its generalization capabilities. The parameterization yields a non-trivial angle both for breaking our overall construction and for improving it. 

Notice, the above definition does not imply that all $\epsilon$-fair models have this property, and some fair models will not be discovered due to that. Empirically, we observed that such models can efficiently be detected and we demonstrate that in the experimental section. 
Additionally, we observe that the output of $\Randomize$ is not required to be indistinguishable from a new sample from $\D$. 
In particular the definition does not rule out that $\Adversary$ is aware of the possible augmentations.

Let $\tilde{T}$ be a sample set $T$ after augmenting each sample. Denote $\emploss{M,\tilde{T}}{g, \Randomize}$ the empirical conditional risk and 
$$EFG = \max_{g_0,g_1 \in \groups}  \left|\emploss{M,\tilde{T}}{g_0, \Randomize} - \emploss{M,\tilde{T}}{g_1, \Randomize}\right|.$$ We state the following, 
\begin{claim}\label{thm:augmented_sampling}
Let  $\M$ be a class of models with $(\epsilon, \alpha, \Randomize)$-detectable fairness. Let $T, \tilde{T}, \Randomize$ and $ M \in \M$ be as stated above. $M$ is $\epsilon$-fair with confidence $1-\delta$ if:
    \begin{align}\nonumber
        EFG < \alpha ~~~~~\text{and}~~~~~
        \min_{g \in \groups}  m_{g} \geq \frac{2}{(\alpha - EFG)^2}\ln{\frac{2|\groups||\Y|}{\delta}} ~~.
    \end{align}
\end{claim}

In other words, we can certify $\epsilon$-fairness of a model with high confidence assuming $(\epsilon, \alpha, \Randomize)$-detectable fairness.
The proof is in the supplementary materials.



\section{Implementing the Framework}
\label{sec:implementation}
We describe how to implement the framework from previous sections using the interactive tests and guarantees from \cref{thm:sampling}. While the implementation is described at a high level, it is easy to instantiate each of the components based on existing cryptographic tools and the experimental results in the Experiments section.

\subsection{Creating a Verification Test}
\label{sec:create_test}
Consider a design of an interactive test based on the set $T=\{(x_1,g_1,y_1),...,(x_m,g_m,y_m)$ and parameters $\delta,\epsilon$ as follows:
\begin{enumerate}
    \item The regulator $\Regulator$ computes the minimal $m_g$ fulfilling \cref{eq:minbound-fairness-private} by assuming $EFG = 0$. If $T$ does not contain enough samples from each group, then $\Regulator$ aborts. If $\Regulator$ does not abort, it tells $\Server$ the total number of inputs $m$ that will be checked.

    \item $\Regulator$ and $\Server$ run a secure computation of a  functionality $\FMcheck$ which is described below. 
    $\Server$ inputs $M$ into $\FMcheck$ while $\Regulator$ inputs $(\{(x_i, g_i, y_i)\}_{i\in [m]})$.
    The functionality $\FMcheck$ consists of the following steps: 
    \begin{enumerate}
        \item Compute $\hat{y}_i \gets M(x_i)$ for all $i\in [m]$.
        \item For all $i\in [m]$, compute a bit $b_i$ as 1 if $\hat{y}_i=y_i$ and 0 otherwise. 
        \item Compute for each group $g$ the empirical risk $\emploss{M,T}{g}$ based on \cref{eq:emp_group_risk} and the $b_i$ values. 
        \item  Evaluate Eq.~(\ref{eq:minbound-fairness-private}) of \cref{thm:sampling} (checking $\epsilon$-fairness). Output $1$ if the statement holds and $0$ otherwise.
        \end{enumerate}
   
\end{enumerate}
Based on the statement of \cref{thm:sampling} it follows that  $\FMcheck$ will output $1$ if and only if the model $M$ provided by $\Server$ is $\epsilon$-fair with confidence $1-\delta$.  

The secure computation of  $\FMcheck$ implements the functionality as a Binary circuit $K$ that is evaluated on secret inputs. 
We examine the size of this circuit in a following section under Efficiency.

\paragraph*{A test using augmented data}
If $\Regulator$ instead wishes to use public and augmented data as for Theorem \ref{thm:augmented_sampling} then this will only work assuming that $M$ is $(\epsilon,\alpha)$-detectable as defined in Definition \ref{def:detectable_fairness}. In such a setting $\Regulator$ would now create a test set $T'$ from $T$ locally using an augmentor $\Randomize$ and then follow the exact same path as for the public data (albeit with different constants). 

\subsection{Algorithms}
\label{sec:algorithms}
We now describe how to use the circuit $\Circuit$ from the previous section to implement the framework. The overall approach is as follows: 
Initially, $\Regulator$ generates a signature key pair and distributes the verification key to all other participants. 
Then $\Regulator$ and $\Server$ run a secure computation which runs the interactive test and computes a Merkle tree hash $\hat{H}_k(M)$ of the model $M$. If the test finds that the model is fair, then $\Regulator$ signs ($\hat{H}_k(M), \epsilon, \delta$, fairness definition string) and sends the signature to $\Server$. By signing $\epsilon, \delta$ and a fairness definition string we allow multiple fairness definitions and hyperparameters to be certified. 

 Later, whenever  $\Server$ and $\Client$ run a certified inference for $\epsilon, \delta$ and a fairness definition, then in addition to running a secure computation of  $M(x)$, 
 the functionality will also recompute the hash $\hat{H}_k(\cdot)$ of the model provided by $\Server$ and output it to $\Client$, while $\Server$ sends the signature on the model to $\Client$. $\Client$ can then locally check if $\Regulator$ originally issued the signature on the hash for those hyperparameters and fairness definition, given the public verification key of $\Regulator$. The overall protocols are outlined in \figref{certified inference}.

\begin{boxfig}{Protocol $\PiFramework$ for Certified Inference}{certified inference}
We will have three participants $\Server,\Client,\Regulator$ as outlined before. Let  $(\Keygen,\Sign,\Verify)$ be a signature scheme and $H_k(\cdot)$ be a collision-resistant hash function whose key $k$ is a common input to all parties. Moreover, let $\FMPC$ be a functionality for secure computation as outlined in \figref{securecomp}. $\Server$ has a model $M$ as input, $\Regulator$ has a fairness validation set $T$ as well as parameters $\epsilon,\delta$ and $fair$ the fairness definition string.
\begin{description}
\item[Setup:] This reflects Step $1$ of Figure \ref{fig:registration_and_verification}.
\begin{enumerate}
    \item $\Regulator$ uses $\Keygen$ to generate a key pair $(sk,vk)$. $\Regulator$ keeps $sk$ private and sends $vk$ to $\Client,\Server$ as $\certificateId$.
\end{enumerate}
\item[Certification:] This reflects Steps $2,3$ of Figure \ref{fig:registration_and_verification}.
\begin{enumerate}
\item $\Server,\Regulator$ input the same values into $\FMcheck$. 
\item Let $\Circuit$ be the circuit as outlined in the previous section. Create a circuit $\Circuit_{\mathtt{cert}}$ that performs the following: 

First run $\Circuit$ on the respective inputs as before, computing $\FMcheck$. Denote the output bit of this circuit as $b$. Then compute the Merkle tree output $h \gets \hat{H}_k(M)$ based on the hash function $H_k$. Finally, output $(b,h)$ to $\Regulator$.
\item Both parties run a  secure computation of  $\Circuit_{\mathtt{cert}}$ using $\FMPC$.
\item 
If the output bit $b$ is 1, then  $\Regulator$ computes $\sigma(h)\gets \Sign_{sk}(h, (\epsilon, \delta, fair))$ and sends it as $\certificateModel$ to $\Server$.
\end{enumerate}
\item[Inference:] This reflects Steps $4,5$ of Figure \ref{fig:registration_and_verification}.
\begin{enumerate}
    \item 
    $\Server$ sends  $\sigma$ to  $\Client$. $\Client$ also knows the signature verification key $vk$.
    
     \item   $\Server$ and  $\Client$ run a secure computation, where $\Server$ inputs $\tilde{M}$  and $\Client$ inputs its input $x$.
    For this secure computation they construct a circuit $\Circuit_{\mathtt{inf}}$ as follows:
    \begin{enumerate}
        \item Compute $\hat{y}\gets \tilde{M}(x)$.
        \item Compute $\tilde{h}\gets \hat{H}_k(\tilde{M})$.
    \end{enumerate}
    \item $\Client,\Server$ run a secure computation of $\Circuit_{\mathtt{inf}}$ using $\FMPC$. $\Client$ obtains as output $(\hat{y},\tilde{h})$ while $\Server$ does not obtain anything. 
    
\item    $\Client$ computes $b\gets \Verify_{vk}(\tilde{\sigma},(\tilde{h}, \epsilon, \delta, fair))$. If this is true then 
 $\Client$ accepts $\hat{y}$. Otherwise it  rejects it.
\end{enumerate}
\end{description}
\end{boxfig}

\subsection{Security}
We provide a sketch of the argument about the security of $\PiFramework$. This must naturally stay on a high level, since we did not make the security properties of the framework formal.

First, we note that $\PiFramework$ leaks to $\Regulator$ and $\Client$ the Merkle-tree hash $h$ of the model. But 
since it can be assumed that $M$ has high entropy and the implementation of $H_k$ is a cryptographic hash function, the leakage of $h$ should be tolerable.\footnote{It is possible  in principle to reduce this leakage by computing  $\Regulator$'s signature of $h$, and the signature verification by $\Client$, in a secure computation, but this will considerably increase the overhead. In the other direction, if we are willing to leak some more information then the circuit $K$ can be modified to output to $\Regulator$ whether $M$ successfully classified each input $x_i$ and let $\Regulator$ compute the $\epsilon$-fairness of the model locally. This will simplify the secure computation at the cost of leaking more data to $\Regulator$.}

That being said, we base our security argument on statements about the security of the building blocks that are used, which can be instantiated using well-known cryptographic constructions:
\begin{itemize}
    \item {\em The functionality $\FMPC$ can be implemented using a secure protocol.} As mentioned above this can be done using secure two-party or multi-party computation (MPC). 
    \item {\em There exist secure signature and hashing schemes.}
\end{itemize}

Given these primitives, we can assume that the certification and inference steps of \figref{certified inference}
 are as secure as if they were computed by a trusted party: Assume that in the inference step, the signature $\tilde{\sigma}$ and output  $\tilde{h}$ of $\FMPC$ are validated. This can only happen due to  3 cases: \begin{enumerate*}[label=(\roman*)] \item $\tilde{\sigma}$ was generated for $\tilde{M}$ by $\Regulator$ (which is the desired course of events); \item $\tilde{\sigma}$ was issued by $\Regulator$ but for a different $\tilde{M}'$; or \item $\tilde{\sigma}$ was never issued by $\Regulator$. \end{enumerate*}
 In the last case, $\Server$ must have broken the security of the signature scheme. In the second case, $\Server$ must have broken the collision-resistance of $H_k$. Therefore, either $\Server$ managed to break the signature scheme or the hash function, or  $\Regulator$  signed $\tilde{M}$. $\Regulator$ computes this signature  if and only if the model passed the interactive test. 
Based on the statement of Theorem \ref{thm:sampling} it follows that this test passes if and only if the model $\tilde{M}$ provided by $\Server$ is $\epsilon$-fair with confidence $1-\delta$. 

\subsection{Efficiency}
\label{sec:efficiency}
We now estimate the efficiency of implementing our framework using $\PiFramework$. 
We first claim that it only makes sense to run our framework in settings where the ML inference is done using a secure computation: If the inference is not computed using a secure computation, then one option is for the client to learn the model and run by itself a check for fairness, or send the model to another party and ask it to do this check. Another option is that the client simply hands over its input to the model owner, but this would require prohibitively expensive zero-knowledge proofs, to be computed at the owner side, to attest to fairness of the output without revealing anything about the model.

Therefore, given that inference is done via secure computation, the parties must incur the cost of running a secure computation of the inference, and the efficiency of the framework should be measured by the additional overhead that is added on top of the secure inference. 

The main computational tasks that are run by $\PiFramework$ are as follows:
\begin{itemize}
    \item The \textbf{Certification} phase runs $m$ instances of a secure computation of inference and in addition computes a hash of the model and checks the accuracy of the output.
    \item The \textbf{Inference} phase runs a single secure computation of the inference and in addition computes a hash of the model.
\end{itemize}
The {\bf Certification} phase is a one-time event, and therefore its overhead is less critical. Theorem~\ref{thm:sampling} shows that the number of samples $m_g$ per group should be $m=\frac{2}{(EFG-\epsilon)^2} \ln \frac{2|G|}{\delta^2}$. Setting for example $EFG=0.05, \epsilon = 0.1, \delta=0.2$ and considering  $|G|=100$ groups, we get that $m_g\approx 6800$, which does not seem to be too far off from existing training set sizes. 

In more detail, we describe here the cost of implementing the different steps of the circuit $K$ which computes the certification:
Step (a) needs to implement the inference $m$ times. This is by far the largest component of the circuit. 
Step (b) computes $m$ comparisons, which are easy. Step (c) computes  $\emploss{M,T}{g}$ for 
each group $g$,  based on Eq.~\ref{eq:emp_group_risk}. This computation must sum the $b$ values for each group $g$. To make this step efficient, the circuit must hard-wire the connections for these summations, and the locations of  the inputs from each $g$ can be known. (There is no need to hide these locations from $\Server$.) Eq.~\ref{eq:emp_group_risk} also computes a division by $m_{g}$, but there is no need to compute the division and the circuit forwards $m_{g}\cdot \emploss{M,T}{g}$ to the next step.  
  Step (d)  tests  Eq.~\ref{eq:minbound-fairness-private} for each pair of $g_0, g_1$, namely computes $\emploss{M,T}{g_0} - \emploss{M,T}{g_1}$. 
  Since the input to this step is $m_{g_i}\cdot \emploss{M,T}{g_i}$
 then the test in this equation should be changed appropriately (which is straightforward, especially if $m_{g_0} = m_{g_1}$).
 
As for the cost of computing $M(\cdot)$, current secure computation implementations for this task only hide the weights of a DNN but reveal the actual network structure and activation functions. We assume that our secure computation will also only hide the weights as this seems to be a standard assumption. Therefore, we ask what is the additional cost of  hashing this data over the default cost of using the weights in the computation of the model. 

There is a lot of current work on lightweight hashing schemes for usage in zero-knowledge proofs, and it is reasonable to expect that a lot of improvements in this area will be made in the near future. 
As a baseline, we consider the Keccak-F function, which is the basis of the SHA3 standard. That function takes a 1600 bit input and can be implemented by a Boolean circuit of 38,400 AND gates (see \citet{bristol_circuits}), i.e. 24 AND gates per input bit.   
If we use a Merkle tree then the total number of hashes is twice the number of input blocks\footnote{We can improve on that by having the circuit output to $\Client$ the results of the first layer of the Merkle tree, and have $\Client$ locally compute the rest of the tree. For this to work, we will on the other hand have to add random values to each input block to avoid lookup table-based attacks on preimages of $H_k$.}. Therefore, the total  
cost is about 48 AND gates per input bit.

Now, with regards to the secure evaluation of the model (not considering special MPC implementations for secure inference\footnote{This analysis neglects recent works such as e.g. \cite{secure_eval_quantized} that apply to special types of networks only. We believe that the accuracy of the networks such as MobileNets that are used in \cite{secure_eval_quantized} is too low to be of use for fairness testing. }), let us consider a setting where the weights have 32 bit fixed-point values. The cost per each weight (when used in DNN inference) must be at least that of multiplying the weight with either an input or output of a hidden layer and adding all these products together (neglecting the cost of the activation function). Multiplying the weight with a 32 bit value costs $185$ ANDs per input bit, while adding up the result would only require $6$ ANDs per input bit (see \cite{bristol_circuits}), and we therefore take the 
 assumption that the total cost of  the secure computation is $191$ AND gates per bit of the weights (neglecting the activation function). Therefore the fairness verification increases the cost of inference in this model by only about $25\%$. While using optimized implementations for inference will make the additional overhead from hashing larger, we can in practice lower the cost of hashing drastically by exploiting special properties of $\FMPC$ which allow the use of homomorphic commitments. We leave such specialized hashing techniques as interesting future work.






\section{Experiments}
\label{sec:experiments}

We provide empirical evidence to demonstrate that the assumptions made for the fairness tests are meaningful. 

\begin{table*}[t!]
        \caption{
        \small
        Fairness test in the private and public settings on the 3 image datasets. "Regular" refers to the model trained fairly, while "Bias" refers to the biased sampled training. \label{tbl:private_public_test}}
    \centering
     \scalebox{0.99}{\begin{tabular}{l  l | c  c | c  c | c  c | c  c} 
     \hline
     \multirow{3}{*}{Dataset} & \multirow{3}{*}{Model} & \multicolumn{4}{c |}{\textbf{Private Setting}} & \multicolumn{4}{c}{\textbf{Public Setting}}\\
     {} & {} & \multicolumn{2}{ c |}{Accuracy} & \multicolumn{2}{ c |}{ORE EFG} & \multicolumn{2}{c | }{Accuracy} & \multicolumn{2}{c}{ORE EFG}\\
     {} & {} & Regular & Bias & Regular & Bias & Regular & Bias & Regular & Bias\\  
     \hline
     UTKFace & ResNet18 & 89.76 & 88.56 & 0.012 & 0.093 & 96.11 & 91.44 & 0.027 & 0.139 \\
     \hline
     C-MNIST & LeNet & 98.11 & 74.01 & 0.001 & 0.450 & 89.17 & 67.97 & 0.007 & 0.340 \\ 
     \hline
     CelebA & ResNet18 & 97.63 & 96.95 & 0.007 & 0.034 & 96.60 & 97.02 & 0.010 & 0.033 \\
     \hline
    \end{tabular}}

\end{table*}



We used six different datasets from various domains: visual (UTKFaces~\cite{utk}, LFW~\cite{LFW}, Colored-MNIST~\cite{arjovsky2019invariant}, and a subset of CelebA\footnote{We annotated 8,500 celebrities out of 10,177 in the dataset for ethnicity using Amazon Mechanical Turk. Three turkers annotated three images of each of the 8,500 celebrities, resulting in 177,683 images. The annotations can be downloaded from \url{www.github.com/will/be/published/}.} \cite{celeba}), tabular (Adult Income~\cite{adult_income}) and spoken (TIMIT~\cite{timit}). The datasets vary in size and disparity of minority groups and as such some can be used to create fair or unfair models based on their empirical fairness gap (EFG). 
We demonstrate the variety of our datasets and detail the preprocess in supplementary materials.
\paragraph{Private Data Setup.} 
\label{sec:exp_private_setup}
In the following setup, we assume that $\Regulator$ possesses a subset of secret samples to be used to certify a model $M$ for fairness and accuracy. Naturally, we split the data into a training and test subset. Setting  $\epsilon$-fair and $\delta$-confidence thresholds, we can certify whether a model is fair using the conditions in $\cref{thm:sampling}$. 
A bottleneck of these conditions is our dependency on the size of the sample set. Datasets with bigger sample set allow us to certify more (fair) models, while we were not able to certify a (fair) model if the sample set was too small, even if it is indeed truly fair under the chosen fairness metric. 

We performed our test on the mentioned datasets with $\delta = 0.05$ and $\epsilon \in \{0.05, 0.075, 0.1\}$. For some tasks this gap and confidence level might be intolerable, but for others, such as gender prediction of a face image, which is the task set for UTKFace, LFW and CelebA, it is better than the existing empirical gaps between ethnicity groups of well-known service providers' models~\cite{buolamwini2018gender}.

\begin{figure}[t!]
    \centering
    \includegraphics[width=0.45\textwidth]{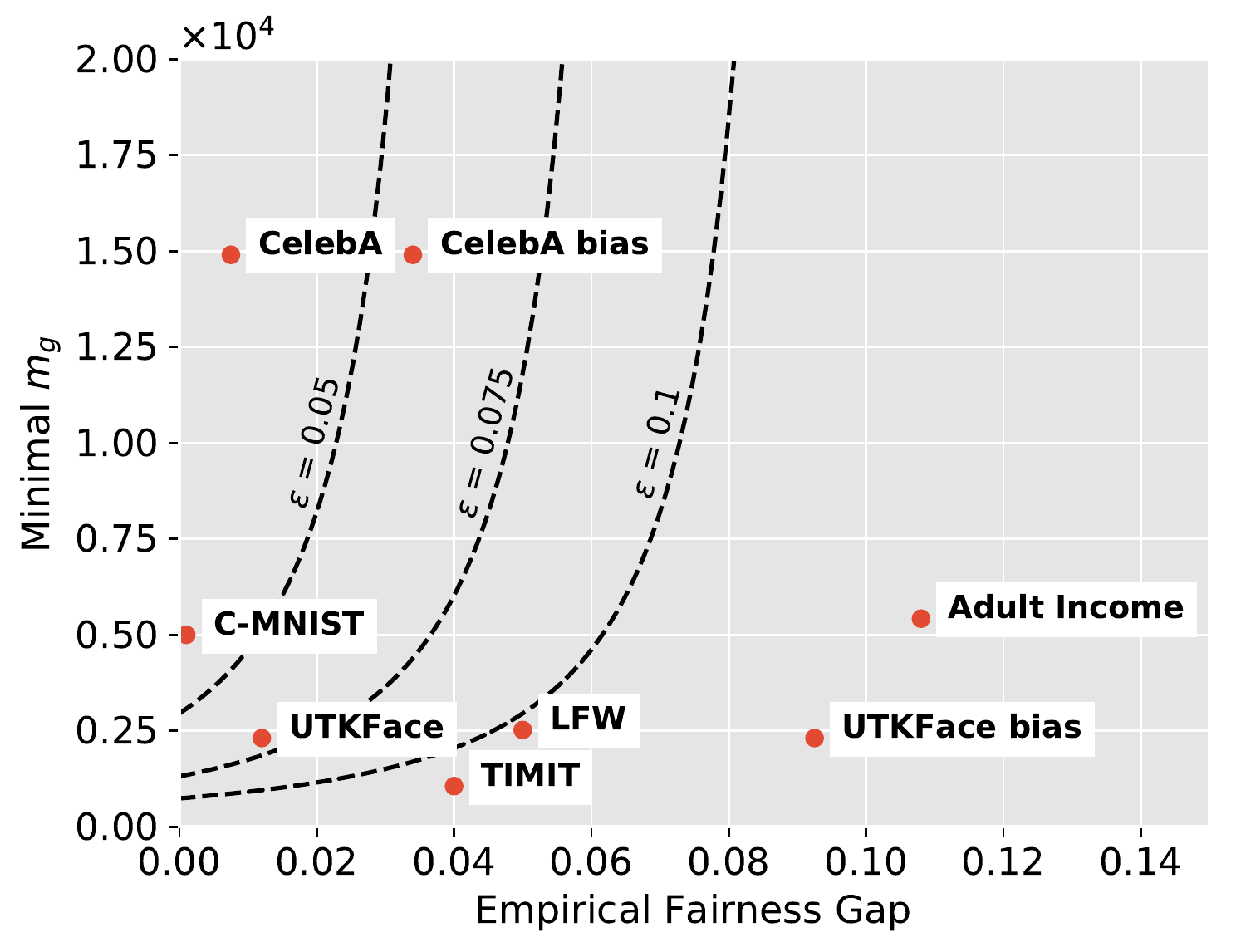}
    \caption{\small Private fairness test borders by EFG and the minimal $m_g$. Left to the dashed border is the area where a model would pass the test for that $\epsilon$ with $\delta = 0.05$. Dots indicate the ORE results for each dataset.}
    \label{fig:private_exp_summary}
\end{figure}

The test results for ORE are shown in Figure~\ref{fig:private_exp_summary}. As shown, out of the six datasets only C-MNIST and CelebA produced fair models during our training for $\epsilon=0.05$, while UTKFace has a fair model for $\epsilon=0.075$. LFW, Adult Income and TIMIT datasets are all below the threshold of all tests, either due to sample size or a large EFG. Therefore, we focus on the first three datasets as they are the only ones to pass any of our tests. Note that by adjusting the allowed bias, $\epsilon$, we can certify the other datasets. For example, the minority group in LFW has only 559 samples. With its current empirical gap, $EFG = 0.049$, choosing $\epsilon = 0.2$ would suffice to certify the LFW model.

\begin{figure*}[t!]
    \centering
    \includegraphics[width=0.9\textwidth]{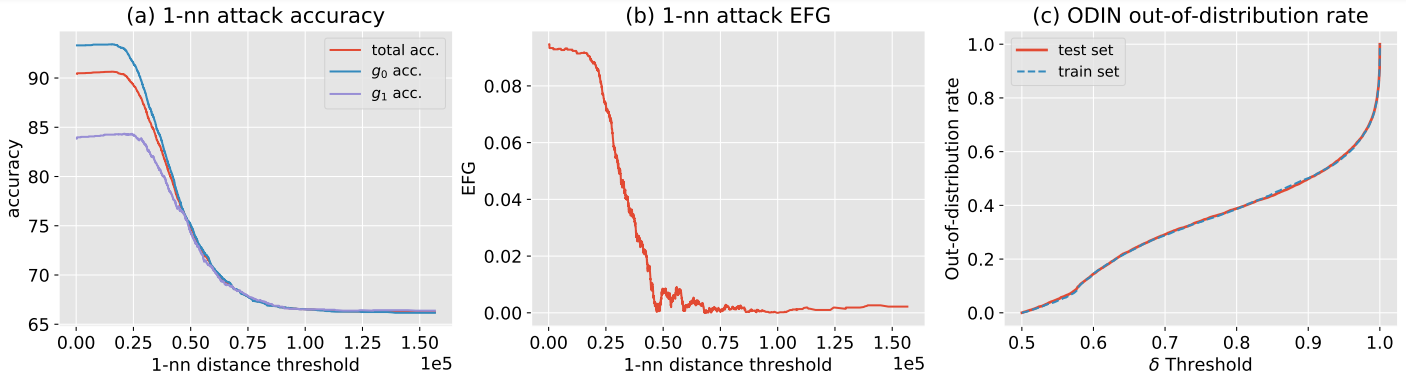}
    \caption{
    \small
    (a)-(b) 1-NN attack accuracy and EFG; (c) ODIN out-of-distribution rate. \label{fig:knn_odin_public_exp}} 
\end{figure*}

To further evaluate the setup, we trained the same models with a tainted batch sampler. The sampler showed less samples from the smallest minority group-label pair $(g_1,y_1)$ in each batch in order to generate a synthetic sample disparity. We denoted these models as \emph{Bias} in Fig. \ref{fig:private_exp_summary} and \cref{tbl:private_public_test}. The taint resulted in an almost as accurate model with a much larger EFG, suggesting they are less fair.
For EO, only CelebA had enough samples to certify a model for $\epsilon=0.05$. It requires at least twice as many samples (since we count $m_{g,y}$ instead of $m_g$). We find it interesting as it implies the amount of data should be a consideration even for which definition of fairness is practical to choose.
We detail the results for EO and DP metrics in the supplementary materials. Results for bias C-MNIST does not appear in Figure~\ref{fig:private_exp_summary} since its performance is significantly worse.

\paragraph{Augmented Public Data Setup.} 
\label{sec:public_augment_exp}
In this setup, all the data  used in the test is known to all participants. With that in mind, we show a potential augmentor for datasets of images and demonstrate empirically that unfair models cannot pass our test as both accurate and fair. Even though the training set is the same as the test set, the key difference is that $\Server$ fixes $M$ before we apply our augmentor to the dataset with new randomness. This generates diverse enough samples, for which models that are fair and generalize well on augmented samples pass the test, while models which are either unfair or bad at generalization fail the test.
Our augmentations include rotation, cropping, blanked pixels \cite{zhong2017random} and added Gaussian noise. Each augmentation was set to be invoked at a certain probability threshold which was chosen randomly. The augmented images should keep the same label and group as the original image to the human eye. By doing so, we hope to generate varied data that cannot be easily reversed or overfitted on.


We tested our three image datasets using the ORE metric, the results are in \cref{tbl:private_public_test}. We used the same method to generate fair and unfair models as in the private data section, except that during training we used the augmentor per sample to generate a new augmented sample each time. When we trained the models on the original dataset, the models were not able to generalize on the augmented data. 

The results show that there exists a margin in EFG between the fair and unfair models on UTKFace, C-MNIST and CelebA, while the margin is different between datasets, potentially due to their varying size and different complexities of the tasks. This suggests the existence of some $\alpha$ per dataset, based on \cref{def:detectable_fairness}, but we were not able to pinpoint the exact $\alpha$. We conducted further attempts to characterize $\alpha$ in supplementary materials. 

\paragraph{Attacks Against Public Fairness Tests.}
As we assume that our test works without knowledge of the concrete model, our scheme might be susceptible to an indirect attack on our augmentor. For example, if the model could distinguish between the public data available and a new sample, it could try and behave fairly during the test, but unfairly when an actual new sample is shown. To mimic such an attack, we tested on UTKFace whether it is easy to fool our test using a simple $k$-nearest neighbour algorithm (kNN) or an out-of-distribution detection technique , ODIN~\cite{odin_liang2017enhancing}, on top of a fair classifier to identify the augmented samples. For a fixed threshold distance from our augmented dataset, we switch to the unfair model and otherwise output the class identified by the kNN. For the ODIN attack, we create a threshold to detect out-of-distribution samples to switch to the unfair classifier. Ideally these attacks use a fair classifier to pass the test as fair when needed, while future new samples (being ``far enough'' in threshold terms) invoke the unfair model as predictor. We gave the kNN augmented samples of the test as referenced neighbors and plotted the accuracy and EFG by the threshold distance for $k=1$ in Fig. \ref{fig:knn_odin_public_exp}a-b (larger $k$ had worse results). In order to have a similar EFG to the fair model, it has to suffer a drop in accuracy; from 91.44\%  to 81.9\%. 

In the other attack, we tuned ODIN's hyperparameters, taking the values which had at least 95\% success rate identifying test samples and had the best results at detecting new samples as out-of-distribution. Further details on tuning are listed in the supplementary materials. We plotted the train and test detection rate as out-of-distribution by threshold in Fig. \ref{fig:knn_odin_public_exp}c. As can be seen, the sets are detected at similar rates, and are nearly indistinguishable. This resulted in a similar fair or unfair behavior depending on the chosen threshold. 

These experiments suggest that these types of attacks are not a good approach to attack the proposed verification test, as this hybrid models cannot pass as both fair and accurate enough for practical applications.

\section{Discussion \& Future Work}
\label{sec:discussion}

We present an interactive test to verify fairness of any machine learning model using cryptographic tools. The interactive test ensures $\Regulator$ does not learn $M$, $x$ does not leak to $\Server$, and $M$ does not leak to $\Client$, yet it verifies the model was used during inference has been certified by $\Regulator$. We experimented with two scenarios where the test data is either public or private. We provide analysis and guarantees for the test data, as well as rigorously define the relation between the empirical fairness gap to the sample set sizes. 

Moreover, from our guarantees and experiments we noticed not all fairness definitions are created equally, some are harder to verify and require a much larger volumes of data, i.e. EO requires at least twice as many samples as ORE. This leaves room for consideration on what practical definition should we aim for with respect to limited resources or what compromise needs to be made in terms of fairness gap and certainty ($\epsilon$ and $\delta$). 

For future work we would like to further explore the public data scenario. Specifically, to characterize the detectable fairness hyper-parameter $\alpha$ and its relation to other parameters like the sample set size $T$, the amount of randomness used per augmentation, etc. Additionally, we would like to explore whether these parameters can be estimated \emph{in advance}, without having to conduct experiments on a dataset. 
Results suggest that this is a challenge on its own. 
Moreover, as we are dealing with large models we also require to hash the model inside secure computation. This step has substantial cost, and it is an open question if it could be made more efficient in practice using different ideas than ours. Lastly, the proposed method is focused on group-based fairness definitions, exploring other fairness definitions is also an interesting research direction.

\bibliographystyle{ACM-Reference-Format}
\balance
\bibliography{./bibs/ml,./bibs/additional}

\newpage
\onecolumn
\appendix
\section{Claims Proofs}
\subsection{\cref{thm:sampling} proof}
\label{proof:thm1}
Using Hoeffding's concentration bound for any $g \in G$:
\begin{align*}
    &\Pr\left[\left|\emploss{M,T}{g} - \loss{M}{g}\right| > \frac{\epsilon - EFG}{2}\right]  \\
     & \leq 2\exp\left(-m_g\frac{(\epsilon - EFG)^2}{2}\right)\\
     & \leq 2\exp\left(-\ln{\frac{2|\groups||\Y|}{\delta}}\right) = \frac{\delta}{|\groups||\Y|}
\end{align*}
Consider a ``good" event where $|\emploss{M,T}{g} - \loss{M}{g}|\leq {(\epsilon - EFG)}/{2}$ for all $g \in \groups$. By a union bound on the complementary event, the probability of this event is at least probability $1- \delta$:
\begin{align*}
    &\Pr\left[\exists g \in G: \left|\emploss{M,T}{g} - \loss{M}{g}\right| > \frac{\epsilon - EFG}{2}\right]
    \\
    &\leq \sum_g \Pr\left[\left|\emploss{M,T}{g} - \loss{M}{g}\right| > \frac{\epsilon - EFG}{2}\right] 
    \\
    & \leq \sum_g \frac{\delta}{|\groups||\Y|} \leq \delta
\end{align*}
Given that the good event holds then, by applying the triangle inequality twice, for any $g_0, g_1 \in \groups$ we have:
\begin{align*}
	|\loss{M}{g_0} - \loss{M}{g_1}| \leq & |\loss{M}{g_0} - \emploss{M,T}{g_0}|   \\
	+ & EFG + |\emploss{M,T}{g_1} - \loss{M}{g_1}| \leq \epsilon
\end{align*}
Hence $\max\nolimits_{g_0,g_1 \in \groups} |\loss{M}{g_0} - \loss{M}{g_1}| \leq \epsilon$ with confidence $1-\delta$. Similar arguments show the same is true for equalized odds and demographic parity.

\subsection{\cref{thm:augmented_sampling} proof}
\label{proof:aug_sampling}
Similar to the proof for\cref{thm:sampling}, we get that:
$$\Pr\left[\left|\emploss{M,\tilde{T}}{g, \Randomize} - \loss{M}{g, \Randomize}\right| > \frac{\alpha - EFG}{2}\right] \leq \frac{\delta}{|\groups||\Y|}$$
which implies that  $|\loss{M}{g_0, \Randomize} - \loss{M}{g_1, \Randomize}| \leq \alpha$ with confidence $1-\delta$ for all groups. Since $\M$ is $(\epsilon, \alpha)$-detectable fairness, then $M$ is $\epsilon$-fair with the same confidence. 
\label{appendix:private_exp}
\begin{table*}[h!]
    \centering
     \begin{tabular}{l  l  c  c  l  c  l  c  l} 
     \hline
     \multirow{2}{*}{Dataset} & \multirow{2}{*}{Model} & \multirow{2}{*}{Accuracy} & \multicolumn{2}{c}{Risk Equality} & \multicolumn{2}{c}{Equalized Odds} & \multicolumn{2}{c}{Demographic Parity} 
     \\ [0.5ex]
     {} & {} & {} & EFG & $\epsilon$-Test & EFG & $\epsilon$-Test & EFG & $\epsilon$-Test 
     \\ 
     \hline\hline
     UTKFace & ResNet18 & 89.76 & 0.012 & Failed$^\dagger$ & 0.067 & Failed & 0.007 & Failed$^\dagger$ \\
     \hline
     UTKFace & Bias-ResNet18 & 88.56 & 0.093 & Failed & 0.115 &  Failed & 0.088 & Failed\\ 
     \hline
     CelebA & ResNet18 & 97.63 & 0.007 & Passed & 0.017 & Passed & 0.083 & Failed\\
     \hline
     CelebA & Bias-ResNet18 & 96.95 & 0.034 & Failed & 0.045 & Failed & 0.039 & Failed\\
     \hline
     C-MNIST & LeNet & 98.11 & 0.001 & Passed & 0.022 & Failed$^\dagger$ & 0.001 & Passed\\
     \hline
     C-MNIST & Bias-LeNet & 74.01 & 0.450 & Failed & 0.485 & Failed & 0.464 & Failed\\
     \hline
     LFW & ResNet18 & 91.06 & 0.049  & Failed$^\dagger$ & 0.398 & Failed & 0.065 & Failed\\
     \hline
     Adult Income & MLP & 84.17 & 0.108 & Failed & 0.377 & Failed &  0.182 & Failed\\
     \hline
     TIMIT & LeNet & 89.07 & 0.040 & Failed$^\dagger$ & 0.117 & Failed & 0.082 & Failed\\
     \hline
    \end{tabular}
    \caption{Fairness test using private data with $\epsilon=0.05$ and $\delta=0.05$.
    ``Failed$^\dagger$'' refers to insufficient sample size to certify the fairness of the model. \label{tbl:private_test}}
\end{table*}

\section{Experiments Details}
\subsection{Datasets}
\label{appendix:datasets}
\begin{itemize}
    \item \textbf{UTKFace} \cite{utk} is a dataset of face images with attribute annotation for age, ethnicity (called \emph{race} and annotated as \emph{black, asian, white} or \emph{other}), and gender (\emph{male} or \emph{female}). We focused on gender prediction as a task across two ethnicity groups \emph{black} and \emph{white} and discarded all other samples, so we were left with 14,604 samples, which we split equally to train and test sets. The dataset consists of 70\% \emph{white} and 30\% \emph{black}, 53\% \emph{male} and 47\% \emph{female}.
    \item \textbf{MNIST} is originally a dataset of hand-written digits from 0 to 9. The dataset is used to predict the digit in the image without additional annotations, hence there are 10 classes across the dataset without any allocation of groups. We changed the task to a binary classification and synthetically generated two fairness groups of digits based on MNIST data. Therefore, we assign the label 0 to the digits 0-4 and the label 1 to the digits 5-9. We randomly colored half of the dataset's digits in red as was done in \cite{arjovsky2019invariant}, resulting in 50\% red digits and 50\% white digits -- these were the fairness groups. We called this dataset \emph{C(olored)-MNIST}.
    \item \textbf{LFW} \cite{LFW} is a dataset of face images with attributes annotation \cite{LFW_attr}. Using the ``Black'' attribute we divided the data into two groups, while using ``Male'' as a binary label.
    \item \textbf{CelebA} \cite{celeba} is a face recognition dataset consisting of more than 10,000 different celebrities with gender labelling. We annotated 8,500 celebrities out of 10,177 in the dataset for ethnicity using Amazon Mechanical Turk. Three turks annotated three images of each of the 8,500 celebrities, resulting in 177,683 images classified as either Asian, African, Caucasian or Other\footnote{The annotations can be downloaded from \url{https://github.com/ShaharKSegal/CelebA_Samples}}. During our experiments we merged all but the Caucasian group to produce a large dataset to showcase our setup, having over 30,000 samples for the minority group.
    \item \textbf{Adult Income} \cite{adult_income} is a tabular features dataset with a label for low/high income. We used the gender feature as group affiliation and income for labels. During the preprocessing all numeric features were normalized, while categorical features were transformed into one-hot vectors in order to be used later by DNN models.
    \item \textbf{TIMIT} \cite{timit} is a voice recognition dataset with dialects and gender annotation. We used the different dialects as groups and gender of speaker as label. To have more samples per dialect, we merged dialects which have much in common and are considered similar, namely we merge New England with New York City and Northern with North Midland, while discarding the rest.
\end{itemize}

\begin{table}[t!]
    \centering
    \begin{tabular}{c c c c c c} 
     \hline
     Dataset & Size & $g_0$, $y_0$ & $g_0$, $y_1$ & $g_1$, $y_0$ & $g_1$, $y_1$\\ 
     \hline\hline
     UTKFace & 14,604 & 37.5\% & 31.51\% & 15.87\% & 15.12\%\\ 
     \hline
     LFW & 13,144 & 74.22\% & 21.52\% & 3.24\% & 1.02\%\\ 
     \hline
     CelebA & 177,683 & 33.97\% & 49.26\% & 8.67\% & 8.11\%\\ 
     \hline
     C-MNIST & 70,000 & 25\% & 25\% &25\% & 25\%\\ 
     \hline
     Adult Income & 48,842 & 46.54\% & 20.3\% &29.52\% & 3.62\%\\ 
     \hline
    \end{tabular}
    \caption{Data distribution across groups and labels for each of the datasets. \label{tbl:data_gender_race} }
\end{table}

\subsection{Private and Public Data Setup Full Results}

The full results for overall risk equality, equalized odds and demographic parity in the private setup can be found in \cref{tbl:private_test}. In the public setup We've also experimented with cutting the UTKFace dataset in half, to see how it affects the margin and $\alpha$. We include results for the LFW dataset. Since we could not generate a fair model in LFW, we have no reference or evidence of margin, but empirically the EFG seems high suggesting the augmentation would work on it as well. The public setup results are in \cref{tbl:public_test}.

\begin{table}[h!]
    \centering
     \begin{tabular}{l  l  c  c  c  c} 
     \hline
     \multirow{2}{*}{Dataset} & \multirow{2}{*}{Model} & \multicolumn{2}{c}{Accuracy} & \multicolumn{2}{c}{Risk Equality EFG} \\
     {} & {} & Fair & Bias & Fair & Bias\\  
     \hline\hline
     UTKFace - Half Size & ResNet18 & 92.13 & 87.03 & 3.56 & 12.91 \\ 
     \hline
     UTKFace & ResNet18 & 96.11 & 91.44 & 2.72 & 13.88 \\ 
     \hline
     C-MNIST & LeNet & 89.17 & 67.97 & 0.65 & 34.04 \\ 
     \hline
     LFW & ResNet18 & 91.98 & - & 7.45 & - \\
     \hline
     CelebA & ResNet18 & 96.60 & 97.02 & 1.01 & 3.28 \\
     \hline
    \end{tabular}
    \caption{Fairness test using public data with an augmentor on the 4 image datasets. "Half Size" refers to the dataset with half of the samples removed. \label{tbl:public_test}}
\end{table}

\subsection{ODIN Tuning}
\label{appendix:odin}
We tuned ODIN's 3 hyperparameters - T temperature, $\epsilon$ perturbation and $\delta$ threshold. We did so in a similar fashion of its original paper~\cite{odin_liang2017enhancing}. We chose T from among $\{1, 10, 100, 1000\}$, $\epsilon$ from 30 evenly spaced numbers between 0 and 0.01 and took those which yielded the best results for any $\delta \in [0,1]$. We note that the hyperparameter tuning had little effect, as most of the values chosen performed very similarly.

\subsection{Testing with unknown margin}
\label{appendix:unknown_alpha}
In certain scenarios it might be hard to determine $\alpha$ necessary for the fairness test in advance. For example, UTKFace in the augmented public data setup has a different fairness gap than the one seen when private data is used, and the fairness gap is influenced by the sample set size. To further investigate the nature of our augmentation under the assumption that the gap is unknown, we tested the models under an increasingly larger degree of augmentation (frequency that each augmentation is invoked) for each sample. The changes in accuracy and fairness gap are presented in \cref{tbl:aug_prob_test}.  The accuracy decreases as we increase the augmentation degree, which fits the idea that it might be hard to generalize on the augmented data. Hence, more augmentation yields less accuracy. The fairness gap, on the other hand, had inconclusive results: increasing the augmentation degree had little or no effect on CelebA dataset, while it greatly varied on UTKFace between fair and unfair models.

 \begin{table}[ht]
    \centering
     \begin{tabular}{c  l  c  c  c c} 
     \hline
     Augmentation & \multirow{2}{*}{Dataset} & \multicolumn{2}{c}{Fair Model} & \multicolumn{2}{c}{Unfair Model} \\ [0.5ex]
     Degree & {} & Acc. & Risk-Eq EFG & Acc. & Risk-Eq EFG \\ 
     \hline \hline
     25\% & UTKFace & 97.08 & 3.56 & 91.74 & 18.51\\ 
     \hline
     50\% & UTKFace & 94.58 & 2.03 & 91.14 & 15.4\\ 
     \hline
     75\% & UTKFace & 91.74 & 3.15 & 88.16 & 13.36\\ 
     \hline
     25\% & CelebA & 97.24 & 1.03 & 96.99 & 4.82\\ 
     \hline
     50\% & CelebA & 95.84 & 0.84 & 96.08 & 4.33\\ 
     \hline
     75\% & CelebA & 93.84 & 0.79 & 94.67 & 4.42\\ 
     \hline
    \end{tabular}
    \caption{Accuracy and overall risk equality EFG for different degrees of augmentation \label{tbl:aug_prob_test}}
\end{table}

\end{document}